\documentclass{IEEEtran}
\usepackage{amsmath}
\usepackage{amssymb}
\usepackage{graphicx}
\usepackage{subcaption}
\usepackage{algorithm}
\usepackage{algorithmic}
\usepackage{multirow}
\usepackage{authblk}

\title{Evolutionary Multitasking for Single-objective Continuous Optimization: Benchmark Problems, Performance Metric, and Baseline Results}
\author[1]{Bingshui Da}
\author[1]{Yew-Soon Ong}
\author[2]{Liang Feng}
\author[3]{A.K. Qin}
\author[1]{Abhishek Gupta}
\author[4]{\\Zexuan Zhu}
\author[5]{Chuan-Kang Ting}
\author[6]{Ke Tang}
\author[7]{Xin Yao}
\affil[1]{School of Computer Science and Engineering, Nanyang Technological University}
\affil[2]{College of Computer Science, Chongqing University}
\affil[3]{Computer Science and Information Technology, RMIT University}
\affil[4]{College of Computer Science and Software Engineering, Shenzhen University}
\affil[5]{Department of Computer Science and Information Engineering, National Chung Cheng University}
\affil[6]{School of Computer Science and Technology, University of Science and Technology of China}
\affil[7]{School of Computer Science, University of Birmingham}

\begin{document}
\maketitle

\section{Introduction}
The notion of \emph{evolutionary multitasking} has recently emerged as a promising approach for automatically exploiting the latent synergies between distinct (but possibly similar) optimization problems, simply by solving them together in a unified representation space \cite{MFO}. Inspired by human's cognitive ability to multitask, evolutionary multitasking aims to improve convergence characteristics across multiple optimization problems at once by seamlessly transferring knowledge between them. A particular mode of knowledge transfer that has  recently been studied in \cite{MFO} and \cite{MOMFO} is that of \emph{implicit genetic transfer} through chromosomal crossover \cite{sudoku}. While the concept of multitask optimization may seem to bear similarity with that of multitask learning \cite{MTL}, it is noted that the former focuses on exploiting shared knowledge for improved \emph{problem-solving}, in contrast to merely learning.

One of the key practical motivations behind our proposition is to effectively deal with the volume, velocity, and complexity of present-day challenges faced in various industrial environments. In particular, multitasking provides a means to \emph{enhance productivity}. To elaborate, in many real-world settings, it is common for problems with essentially similar properties to recur. In such cases, it makes little sense to re-explore the same design space repeatedly from scratch via \emph{tabula rasa} optimization techniques. Herein, evolutionary multitasking enables autonomous transfer of knowledge from one problem to the next, thereby naturally augmenting the effectiveness of the search.
	
In addition to enhancing productivity, the process of multitasking can often provide superior quality solutions in comparison to traditional single-task optimization. This is most often true in cases when the fitness landscapes of the constitutive tasks are appropriately complementary. A first step toward quantifying the synergy (or correlation) between optimization problems (with known fitness landscapes) has recently been presented in \cite{FSM}. It is contended that investigating the correspondence between the measured synergy between problems, and the performance of a multitasking engine, can provide necessary insights into the mechanisms of various genetic operators and how they may be enhanced to better suit the practice of evolutionary multitasking in the future. To this end, the present technical report provides a diverse set of benchmark problems, with associated performance metric and baseline results, to support the future development of the field.

In \cite{FSM}, a metric is proposed for quantifying inter-task correlations between tasks, which involves the calculation of partial derivatives and integrals over the entire search space (and can therefore be time consuming to compute). Accordingly, in this technical report, we propose a simpler derivative/integration-free alternative for rapidly computing inter-task synergies, one that is based on Spearman's rank correlation coefficient (as is described in Section IV). This metric is then used to quantify the synergy between variants of popularly used single-task optimization functions. Based on the aforestated results, a diverse set of nine problem combinations are constructed, that serve as the first set of composite benchmark problems for evolutionary multitasking. A detailed description of the benchmark problem specifications are provided in Section V of this document, with MATLAB implementations of the same made available as supplementary material. Further, sample baseline results have also been provided herein for all the examples.


\section{Formalization of Evolutionary Multitasking}
The concept of evolutionary multitasking has only recently been formalized in \cite{MFO} under the label of \emph{multifactorial optimization} (MFO), where each constitutive task is considered to impart an additional influence on the evolutionary process of a single population of individuals. In this section, we briefly introduce the associated \emph{multifactorial evolutionary algorithm} (MFEA) employed to generate baseline results in this report - source code of the MFEA is made available as supplementary material. 

\subsection{Preliminary}
Consider a hypothetical situation wherein $K$ self-contained optimization tasks are to be performed concurrently. Without loss of generality, all tasks are assumed to be minimization problems. The $j^{th}$ task, denoted $T_j$, is considered to have a search space $X_j$ on which the objective function is defined as $F_j : \pmb{X}_j \rightarrow \mathbb{R}$. In addition, each task may be constrained by several equality and/or inequality conditions that must be satisfied for a solution to be considered feasible. In such a setting, we define MFO as an evolutionary multitasking paradigm that aims to simultaneously navigate the design space of all tasks, constantly building on the implicit parallelism of population-based search so as to rapidly deduce $\{\pmb{x}_1,\pmb{x}_2,\dots, \pmb{x}_{K-1}, \pmb{x}_K \} = arg min \{F_1(\pmb{x}), F_2(\pmb{x}), \dots, F_{K-1}(\pmb{x}), F_K(\pmb{x})\}$, where $\pmb{x}_j$ is a feasible solution in $\pmb{X}_j$. As suggested by the nomenclature, herein each $F_j$ is treated as an additional factor influencing the evolution of a single population of individuals. For this reason, the composite problem may also be referred to as a $K$-factorial problem. 

While designing evolutionary solvers for MFO, it is necessary to formulate a general technique for comparing population members in a multitasking environment. To this end, we first define a set of properties for every individual $p_i$, where $i \in \{1, 2, …, |P|\}$, in a population $P$. Note that the individuals are encoded in a unified search space $\pmb{Y}$ encompassing $\pmb{X}_1, \pmb{X}_2, \dots, \pmb{X}_K$, and can be decoded into a task-specific solution representation with respect to each of the $K$ optimization tasks. The decoded form of $p_i$ can thus be written as $\{\pmb{x}_{i1}, \pmb{x}_{i2}, \dots, \pmb{x}_{iK}\}$, where $\pmb{x}_{i1} \in \pmb{X}_1$, $\pmb{x}_{i2} \in \pmb{X}_2$, $\dots$, and $\pmb{x}_{iK} \in \pmb{X}_K$.

\begin{itemize}
\item \emph{Definition 1}(\emph{Factorial Cost}):  For a given task $T_j$, the \emph{factorial cost} $\Psi_{ij}$ of individual $p_i$ is given by $\Psi_{ij} = \lambda \cdot \delta_{ij} + F_{ij}$; where $\lambda$ is a large penalizing multiplier, $F_{ij}$ and $\delta_{ij}$ are the objective value and the total constraint violation, respectively, of $p_i$ with respect to $T_j$. Accordingly, if $p_i$ is feasible with respect to $T_j$ (zero constraint violation), we have $\Psi_{ij} = F_{ij}$.
\item \emph{Definition 2}(\emph{Factorial Rank}):  The \emph{factorial rank} $r_{ij}$ of $p_i$ on task $T_j$ is simply the index of $p_i$ in the list of population members sorted in ascending order with respect to factorial cost $\Psi_{ij}$.
\end{itemize}

Note that, while assigning factorial ranks, whenever $\Psi_{1j} = \Psi_{2j}$ for a pair of individuals $p_1$ and $p_2$, the parity is resolved by random tie-breaking.

\begin{itemize}
\item \emph{Definition 3}(\emph{Skill Factor}): The \emph{skill factor} $\tau_i$ of $p_i$ is the one task, amongst all other tasks in a $K$-factorial environment, with which the individual is associated. If $p_i$ is evaluated for all $K$ tasks then $\tau_i = argmin_j\{r_{ij}\}$, where $j \in \{1, 2, \dots, K\}$.

\item \emph{Definition 4}(\emph{Scalar Fitness}): The \emph{scalar fitness} of $p_i$ in a multitasking environment is given by $\varphi_i = 1/r_{iT}$, where $T = \tau_i$. Notice that $max\{\varphi_i\} = 1$.
\end{itemize}

Once the fitness of every individual has been scalarized according to Definition 4, performance comparison can then be carried out in a straightforward manner. For example, individual $p_1$ will be considered to dominate individual $p_2$ in multifactorial sense simply if $\varphi_1 > \varphi_2$.

It is important to note that the procedure described heretofore for comparing individuals is not absolute. As the factorial rank of an individual, and implicitly its scalar fitness, depends on the performance of every other individual in the population, the comparison is in fact population dependent. Nevertheless, the procedure guarantees that if an individual $p^*$ attains the global optimum of any task then $\varphi^* = 1$, which implies that $\varphi^* \geq \varphi_i$ for all $i \in \{1, 2, \dots, |P|\}$. Therefore, it can be said that the proposed technique is indeed consistent with the ensuing definition of multifactorial optimality.

\begin{itemize}
\item \emph{Definition 5}(\emph{Multifactorial Optimality}): An individual $p^*$ is considered to be optimum in multifactorial sense if there exists at least one task in the $K$-factorial environment which it globally optimizes.
\end{itemize}

\subsection{Multifactorial Evolution: A Framework for Effective Multitasking}
In this subsection we describe the Multifactorial Evolutionary Algorithm (MFEA), an effective multitasking framework that draws upon the bio-cultural models of multifactorial inheritance \cite{ong}. As the workings of the approach are based on the transmission of biological as well as cultural building blocks from parents to their offspring, the MFEA is regarded as belonging to the realm of \emph{memetic computation} \cite{nineteen, twenty} -- a field that has recently emerged as a successful computational paradigm synthesizing Darwinian principles of natural selection with Dawkins’ notion of a meme as the basic unit of cultural evolution \cite{twentyone}.  An overview of the procedure is provided next.

As shown in Algorithm 1, the MFEA starts by randomly creating a population of $n$ individuals in the unified search space $\pmb{Y}$. Moreover, each individual in the initial population is pre-assigned a specific skill factor (see Definition 3) in a manner that guarantees every task to have fair number of representatives. We would like to emphasize that the skill factor of an individual (i.e., the task with which the individual is associated) is viewed as a computational representation of its pre-assigned cultural trait. The significance of this step is to ensure that an individual is only evaluated with respect to a single task (i.e., only its skill factor) amongst all other tasks in the multitasking environment. Doing so is considered practical since evaluating every individual exhaustively for every task will generally be computationally demanding, especially when $K$ (the number of tasks in the multitasking environment) becomes large. The remainder of the MFEA proceeds similarly to any standard evolutionary procedure. In fact, it must be mentioned here that the underlying genetic mechanisms may be borrowed from any of the plethora of population-based algorithms available in the literature, keeping in mind the properties and requirements of the multitasking problem at hand. The only significant deviation from a traditional approach occurs in terms of offspring evaluation which accounts for cultural traits via individual skill factors.

\begin{algorithm}[t!]
\caption{Pseudocode of the MFEA}
\label{mfea}
\begin{algorithmic}[1]
\STATE Randomly generate $n$ individuals in $\pmb{Y}$ to form initial population $P_0$
\FOR{\textbf{every} $p_j$ in $P_0$}
	\STATE Assign skill factor $\tau_j=\mod(j, K)+1$, for the case of $K$ tasks
	\STATE Evaluate $p_j$ for task $\tau_j$ only
\ENDFOR
\STATE Compute scalar fitness $\varphi_j$ for every $p_j$
\STATE Set $t=0$
\WHILE{stopping conditions are not satisfied}
	\STATE $C_t = \text{Crossover}+\text{Mutate}(P_t)$
	\FOR{\textbf{every} $c_j$ in $C_t$}
		\STATE Determine skill factor $\tau_j$ $\rightarrow$ Refer Algorithm 2
		\STATE Evaluate $c_j$ for task $\tau_j$ only
	\ENDFOR
	\STATE $R_t = C_t \cup P_t$
	\STATE Update scalar fitness of all individuals in $R_t$
	\STATE Select $N$ fittest members from $R_t$ to form $P_{t+1}$
	\STATE Set $t = t+1$
\ENDWHILE
\end{algorithmic}
\end{algorithm}

Following the memetic phenomenon of \emph{vertical cultural transmission} \cite{seventeen, eighteen, nineteen}, offspring in the MFEA experience strong cultural influences from their parents, in addition to inheriting their genes. In gene-culture co-evolutionary theory, vertical cultural transmission is viewed as a mode of inheritance that operates in tandem with genetics, and leads to the phenotype of an offspring being directly influenced by the phenotype of its parents. The algorithmic realization of the aforementioned notion is achieved in the MFEA via a \emph{selective imitation strategy}. In particular, selective imitation is used to mimic the commonly observed phenomenon that offspring tend to imitate the cultural traits (i.e., skill factors) of their parents.  Accordingly, in the MFEA, an offspring is only decoded (from the unified genotype space $\pmb{Y}$ to a task-specific phenotype space) and evaluated with respect to a single task with which at least one of its parents is associated. As has been mentioned earlier, selective evaluation plays a role in managing the computational expense of the MFEA. A summary of the steps involved is provided in Algorithm 2.

\begin{algorithm}[t!]
\caption{Vertical cultural transmission via selective imitation}
\label{vct}
\begin{algorithmic}[1]
\item [Consider offspring $c \in C_t$ where $c = \text{Crossover} $ ]
\item [$+~\text{Mutate}(p_1, p_2)$]
\STATE Generate a random number $rand$ between 0 and 1
\IF{$rand \leq 0.5$ }
	\item[~~~~~~~~~~~ $c$ imitates skill factor of $p_1$]
\ELSE
	\item[~~~~~~~~~~~ $c$ imitates skill factor of $p_2$]
\ENDIF
\end{algorithmic}
\end{algorithm}

\subsection{Search Space Unification}
The core motivation behind the evolutionary multitasking paradigm is the autonomous exploitation of known or latent commonalities and/or complementarities between distinct (but possibly similar) optimization tasks for achieving faster and better convergence characteristics. One of the possible means of harnessing the available synergy, at least from an evolutionary perspective, is through implicit genetic transfer during crossover operations. However, for the relevant knowledge to be transferred across appropriately, i.e., to ensure effective multitasking, it is pivotal to first describe a genotypic unification scheme that suits the requirements of the multitasking problem at hand. In particular, the unification serves as a higher-level abstraction that constitutes a \emph{meme space}, wherein building blocks of encoded knowledge are processed and shared across different optimization tasks. This perspective is much in alignment with the workings of the human brain, where knowledge pertaining to different tasks are abstracted, stored, and re-used for relevant problem solving exercises whenever needed.

Unification implies that genetic building blocks \cite{twentytwo} corresponding to different tasks are contained within a single pool of genetic material, thereby facilitating the MFEA to process them in parallel. To this end, assuming the search space dimensionality of the $j^{th}$ optimization task (in isolation) to be $D_j$, a unified search space $\pmb{Y}$ comprising $K$ (traditionally distinct) tasks may be defined such that $D_{multitask} = max_j\{D_j\}$, where $j \in \{1, 2, \dots, K\}$. In other words, while handling $K$ optimization tasks simultaneously, the chromosome $\pmb{y} \in \pmb{Y}$ of an individual in the MFEA is represented by a vector of $D_{multitask}$ variables. While addressing the $j^{th}$ task, we simply extract $D_j$ variables from the chromosome and decode them into a meaningful solution representation for the underlying optimization task. In most cases, an appropriate selection of $D_j$ task-specific variables from the list of $D_{multitask}$ variables is crucial for the success of multitasking. For instance, if two distinct variables belonging to two different tasks have similar phenotypic (or contextual) meaning, then they should intuitively be associated to the same variable in the unified search space $\pmb{Y}$. On the other hand, in many naive cases where no a priori understanding about the phenotype space is available, simply extracting the \emph{first} $D_j$ variables from the chromosome can oftentimes be a viable option \cite{MFO}.

In what follows, we demonstrate how chromosomes in a unified genotype space can be decoded into meaningful task-specific solution representations when a \emph{random-key unification scheme} \cite{twentythree} is adopted. According to the random-key scheme, each variable of a chromosome is simply encoded by a \emph{continuous value} in the range [0, 1]. The salient feature of this representation is that it elegantly accommodates a wide variety of problems in continuous as well as discrete optimization, thereby laying the foundation for a cross-domain multitasking platform. 
Note that this report only focuses on synthetic continuous optimization problems, thereby decoding can be achieved in a straight-forward manner by linearly mapping each random-key from the genotype space to the box-constrained phenotype space of the relevant optimization task \cite{MFO}. For some guidelines on decoding for discrete optimization problems, the reader is referred to \cite{ong}.

\section{Individual Tasks in Composite Benchmark Problems}
In this technical report, we use 7 commonly used optimization functions as ingredients to form synthetic multitask benchmark problems. The definitions of these individual test functions 
are shown in the following, where $D$ denotes the dimensionality of the search space and $\boldsymbol{x} = [x_1, x_2, \dots, x_D]$ is the design variable vector.
\subsubsection{Sphere} 
\begin{equation}
F_1(\boldsymbol{x}) = \sum_{i=1}^D x_i^2, ~~\boldsymbol{x} \in [-100,100]^D
\end{equation}

\subsubsection{Rosenbrock}
\begin{equation}
\begin{split}
&F_2(\boldsymbol{x}) = \sum_{i=1}^{D-1}(100(x_i^2 - x_{i+1})^2 + (x_i - 1)^2),\\
&~~\boldsymbol{x} \in [-50,50]^D
\end{split}
\end{equation}

\subsubsection{Ackley}
\begin{equation}
\begin{split}
F_3(\boldsymbol{x}) &= -20 \exp\Bigg(-0.2 \sqrt{\frac{1}{D} \sum_{i=1}^D x_i^2}\Bigg) \\
	- \exp\Bigg(\frac{1}{D} \sum_{i=1}^D &\cos(2\pi x_i)\Bigg) + 20 + e,~\boldsymbol{x} \in [-50,50]^D
\end{split}
\end{equation}

\subsubsection{Rastrgin}
\begin{equation}
\begin{split}
&F_4(\boldsymbol{x}) = \sum_{i=1}^D \big(x_i^2 -10\cos(2\pi x_i)+10\big), \\
&~~\boldsymbol{x} \in [-50,50]^D
\end{split}
\end{equation}

\subsubsection{Griewank}
\begin{equation}
\begin{split}
&F_5(\boldsymbol{x}) = 1 + \frac{1}{4000} \sum_{i=1}^D x^2_i - \prod_{i=1}^D \cos\Big(\frac{x_i}{\sqrt{i}}\Big),\\
&~~\boldsymbol{x} \in [-100,100]^D
\end{split}
\end{equation}

\subsubsection{Weierstrass}
\begin{equation}
\begin{split}
F_6(\boldsymbol{x}) &= \sum_{i=1}^D \Bigg(\sum_{k=0}^{k_{\max}}\big[a^k \cos\big(2\pi b^k (x_i +0.5)\big)\big]\Bigg)\\
&- D \sum_{k=0}^{k_{\max}} \big[a^k \cos(2\pi b^k \cdot 0.5)\big]\\
&a=0.5,b=3,k_{\max}=20,~~\boldsymbol{x} \in [-0.5,0.5]^D
\end{split}
\end{equation}

\subsubsection{Schwefel}
\begin{equation}
\begin{split}
&F_7(\boldsymbol{x})  = 418.9829\times D - \sum_{i=1}^D x_i \sin\big(|x_i|^\frac{1}{2}\big),\\
&~~\boldsymbol{x} \in [-500,500]^D
\end{split}
\end{equation}

\begin{table*}[!ht]
\centering
\caption{Summary of properties of Problem Pairs for Evolutionary Multitasking}
	\begin{tabular}{c c c c c}
		\hline 
		Function 	&	Global Minimum ($\pmb{x}^*$)	&	$F(\pmb{x}^*)$	&	Multimodal?	&	Separable?	\\	\hline
		Sphere 		& 	$(0,0,\dots,0)^{\text{T}}$	&	0	&	no	&	yes	\\	\hline
		Rosenbrock 	&	$(1,1,\dots,1)^{\text{T}}$	&	0	&	yes	& 	no	\\ 	\hline
		Ackley  	& 	$(0,0,\dots,0)^{\text{T}}$	&	0	&	yes	&	no	\\	\hline
		Rastrigin	&	$(0,0,\dots,0)^{\text{T}}$	&	0	&	yes	& 	yes	\\	\hline
		Griewank 	&	$(0,0,\dots,0)^{\text{T}}$	&	0	&	yes	& 	no	\\	\hline
		Weierstrass &	$(0,0,\dots,0)^{\text{T}}$	&	0	&	yes	& 	yes	\\	\hline
		Schwefel	&	$(420.9687,\dots,420.9687)^{\text{T}}$	&	0	&	yes	& 	yes	\\	\hline
	\end{tabular}
\end{table*}

\section{Construction of Benchmark Problems}
From the previous studies on evolutionary multitasking \cite{MFO,FSM,ong}, the degree of intersection of the global optima, and the correspondence in the fitness landscape, are two important ingredients that lead to the complementarity between different optimization tasks. To quantify the overall inter-task synergy, the Spearman's rank correlation is proposed herein (as described next).

\subsection{Spearman's rank correlation coefficient}
Consider a large number $n$ of well distributed solutions in a unified representation space. Let the $i^{th}$ solution $\boldsymbol{y}_i$ decode as $\boldsymbol{x}_{i1}$ and $\boldsymbol{x}_{i2}$ in the phenotype space of $T_1$ and $T_2$, respectively, in a multitasking environment. Further, let $r(\boldsymbol{x}_{i1})$ and $r(\boldsymbol{x}_{i2})$ be the \emph{factorial rank} of the  $i^{th}$ solution with respect to the two tasks. Then, the Spearman's rank (or ordinal) correlation coefficient \cite{Spearman} can be computed as the Pearson correlation between $r(\boldsymbol{x}_{i1})$ and $r(\boldsymbol{x}_{i2})$, stated as follows:

$$R_s = \frac{cov(r(\boldsymbol{x}_1),r(\boldsymbol{x}_2))}{std(r(\boldsymbol{x}_1))std(r(\boldsymbol{x}_2))}$$

In our experiments, we randomly generate 1,000,000 points in the unified search space to calculate $R_s$, which is viewed as a representation of the synergy between tasks.

In order to demonstrate the significance of the ordinal correlation measure in evolutionary multitasking, we consider the case of two minimization tasks  $T_1$ and $T_2$ with objective/cost functions $f_1$ and $f_2$ that share high ordinal correlation (i.e., high similarity). To elaborate, we make the extreme assumption that for any pair of solutions $i$, $j$ in the unified search space, we have $f_1(\boldsymbol{x}_{i1})<f_1(\boldsymbol{x}_{j1})\Leftrightarrow f_2(\boldsymbol{x}_{i2})<f_2(\boldsymbol{x}_{j2})$. Thus, it can easily be seen that on bundling these two tasks together in a single multitasking environment, any series of steps leading to a cost reduction for $T_1$ will automatically lead to a cost reduction for $T_2$ \emph{for free}, and vice versa, without the need for any additional function evaluations. In other words, at least within the family of functions characterized by high ordinal correlation, multitask optimization is guaranteed to be effective and provides the scope for \emph{free lunches} \cite{freelunch}. 

\begin{figure*}[!t]
\centering
\includegraphics[width=.8\linewidth]{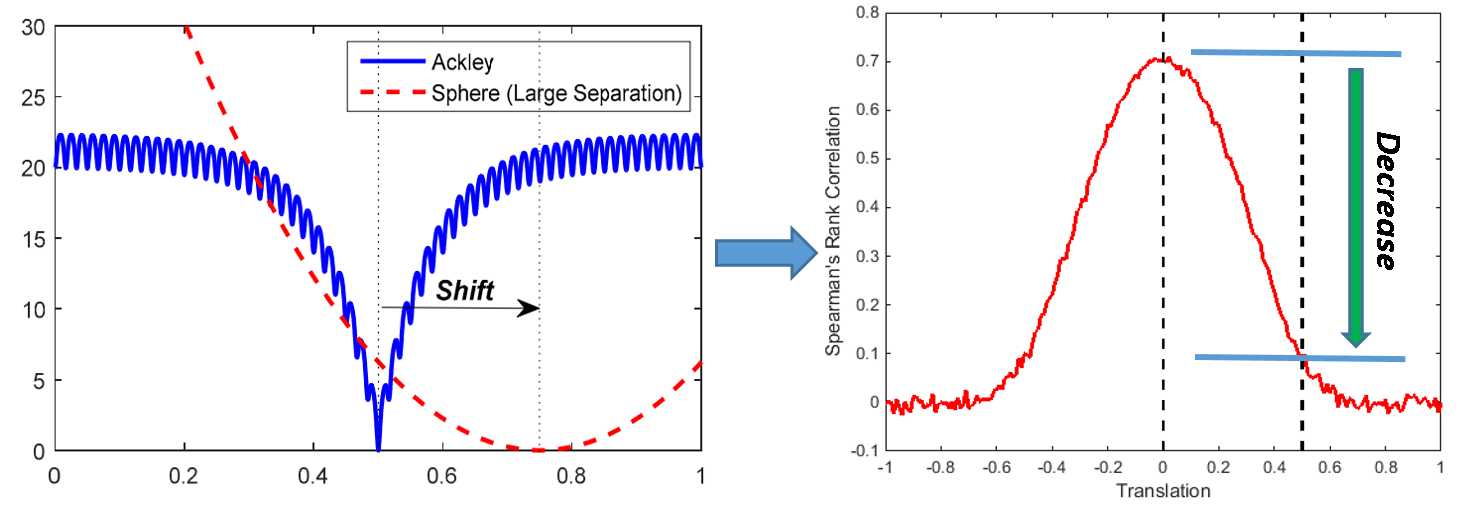}
\caption{The similarity decreases as we shift the Sphere function.}
\label{fig:shift}
\end{figure*}

\subsection{Similarity Tuning and Classification of Benchmark Problems}
Based on the aforementioned similarity metric, it is found the similarity between test functions changes as we shift or rotate the test problems. For example, in Fig. \ref{fig:shift}, we have a 1-D Ackley function and Sphere function (as described in Section III). When we shift the Sphere function, $R_s$ almost monotonically decreases as the distance between the global optima increases. In another aspect, function rotation could generate the alignment and misalignment of the local minimum, thereby providing another means of manipulating the similarity between test problems. To elaborate, in this report, the problem pair with $R_s<0.2$ is considered in low similarity, the problem pair with $0.2\leq R_s < 0.8$ is considered in medium similarity, and the problem pair with $R_s \geq 0.8$ is considered in high similarity. 

We first classify the composite benchmark problems into three categories based on the degree of intersection of the global optima of the constitutive optimization tasks, i.e., complete intersection, partial intersection, and no intersection, as described in Table \ref{tab:intersection}. Note that the intersection of optima occurs in the unified search space, and thus may not necessarily be apparent in the original search space of the optimization tasks. Within each category of optima intersection, there are three sub-categories of inter-task synergy based on the aforestated Spearman's rank correlation similarity metric, i.e., high, medium and low similarity. We design a single problem per category, as a result, nine problem sets are formed in total (as shall be detailed in Section V). 

\begin{table}[h!]
\centering
\caption{Implication of degree of intersection in the unified search space for a pair of optimization tasks}
\label{tab:intersection}
	\begin{tabular}{l p{5cm}}
		\hline
		Category				&	Description	\\	\hline
		Complete intersection	&	The global optima of the two tasks are identical in the unified search space with respect to all variables	\\	\hline
		Partial intersection	&	The global optima of the two tasks are identical in the unified search space with respect to a subset of variables only, and different with respect to the remaining variables	\\	\hline
		No intersection			&	The global optima of the two tasks are different in the unified search space with respect to all variables	\\	\hline
	\end{tabular}
\end{table}

Note that many practical settings give rise to a third condition for categorizing potential multitask optimization settings, namely, based on the phenotypic overlap of the decision variables \cite{ong}. To elaborate, a pair of variables from distinct tasks may bear the same semantic (or contextual) meaning, which leads to the scope of knowledge transfer between them. However, due to the lack of substantial contextual meaning in the case of synthetic benchmark functions, such a condition for describing the overlap/similarity between tasks is not applied in this technical report.

\section{Problem Specifications and Baseline Results}
In this section, we present details of the nine MFO benchmark problems. Baseline results with MFEA and SOEA are also provided (implementations of both solvers are made available as supplementary material). In the current experimental setup, we employ a population of 100 individuals in the MFEA and SOEA. The total number of function evaluations for a composite problem of 2 distinct optimization tasks is restricted to 100,000 (no separate local search steps are performed). Note that ``a function evaluation" here means a calculation of the objective function of a particular task $T_i$, and the function evaluations on different tasks are not distinguished. In MFEA, the random mating probability (\emph{rmp}) is set to 0.3 \cite{MFO}. The simulated binary crossover (SBX) \cite{SBX} and polynomial mutation operators are employed for offspring creation for both MFEA and SOEA. 
Further note that no uniform crossover-like random variable swap between offspring is applied in our experiments (thereby reducing schema disruption), so as to clearly bring out the unadulterated effects of implicit genetic transfer during multitasking. All results presented hereafter are averages of 20 runs of the two solvers. All parameter settings are kept identical for both solvers to ensure fair comparison.

In what follows, $\boldsymbol{M}$ is used to represent rotation matrix (details provided below and in supplementary material). Further, $\boldsymbol{o}$ represents the location of the global optimum of an optimization task in the original (not unified) search space (details below). Table \ref{tab:summary} summarizes the properties of the test problems.

\begin{table*}[!ht]
\centering
\caption{Summary of properties of Problem Pairs for Evolutionary Multitasking}
\label{tab:summary}
	\begin{tabular}{l l l l c}
		\hline 
		Category				&	Task				&	Landscape 		&	Degree of intersection					&	Inter-task similarity $R_s$	\\	\hline
		\multirow{2}{*}{CI+HS}	&	Griewank ($T_1$)		&	multimodal,	nonseparable	&	\multirow{2}{*}{Complete intersection}	&	\multirow{2}{*}{1.0000}		\\	\cline{2-3}
								&	Rastrigin ($T_2$)	& 	multimodal,	nonseparable 	&											&								\\	\hline
		\multirow{2}{*}{CI+MS} 	& 	Ackley ($T_1$)		& 	multimodal,	nonseparable	&	\multirow{2}{*}{Complete intersection}	&	\multirow{2}{*}{0.2261}		\\	\cline{2-3}
								&	Rastrigin ($T_2$)	&	multimodal,	nonseparable	&											&								\\ 	\hline
		\multirow{2}{*}{CI+LS}	& 	Ackley ($T_1$)		&	multimodal,	nonseparable	&	\multirow{2}{*}{Complete intersection}	&	\multirow{2}{*}{0.0002}		\\	\cline{2-3}
								&	Schwefel ($T_2$) 	& 	multimodal,	separable		&											&								\\	\hline
		\multirow{2}{*}{PI+HS}	& 	Rastrigin ($T_1$)	&	multimodal,	nonseparable	&	\multirow{2}{*}{Partial intersection}	&	\multirow{2}{*}{0.8670}		\\	\cline{2-3}
								&	Sphere ($T_2$)		&	unimodal,	separable		&											&								\\	\hline
		\multirow{2}{*}{PI+MS}	&	Ackley ($T_1$)		&	multimodal,	nonseparable	&	\multirow{2}{*}{Partial intersection}	&	\multirow{2}{*}{0.2154}		\\	\cline{2-3}
								&	Rosenbrock ($T_2$)	&	multimodal,	nonseparable	&											&								\\	\hline
		\multirow{2}{*}{PI+LS}	&	Ackley ($T_1$)		&	multimodal,	nonseparable	&	\multirow{2}{*}{Partial intersection}	&	\multirow{2}{*}{0.0725}		\\	\cline{2-3}
								&	Weierstrass ($T_2$)	&	multimodal,	nonseparable 	&											&								\\	\hline
		\multirow{2}{*}{NI+HS}	&	Rosenbrock ($T_1$)	&	multimodal,	nonseparable	&	\multirow{2}{*}{No intersection}		&	\multirow{2}{*}{0.9434}		\\	\cline{2-3}
								&	Rastrigin ($T_2$)	&	multimodal,	nonseparable	&											&								\\	\hline
		\multirow{2}{*}{NI+MS}	&	Griewank ($T_1$)	&	multimodal,	nonseparable	&	\multirow{2}{*}{No intersection}		&	\multirow{2}{*}{0.3669}		\\	\cline{2-3}
								&	Weierstrass ($T_2$)	&	multimodal,	nonseparable	&											&								\\	\hline
		\multirow{2}{*}{NI+LS}	&	Rastrigin ($T_1$)	&	multimodal, nonseparable	& 	\multirow{2}{*}{No intersection}		&	\multirow{2}{*}{0.0016}		\\	\cline{2-3}
								&	Schwefel ($T_2$)	&	multimodal, separable		&											&								\\	\hline
	\end{tabular}
\end{table*}

\subsection{Benchmark Problems with Complete Intersection of Global Optima}
\begin{figure}
\centering
\begin{subfigure}{.25\textwidth}
  \centering
  \includegraphics[width=\linewidth]{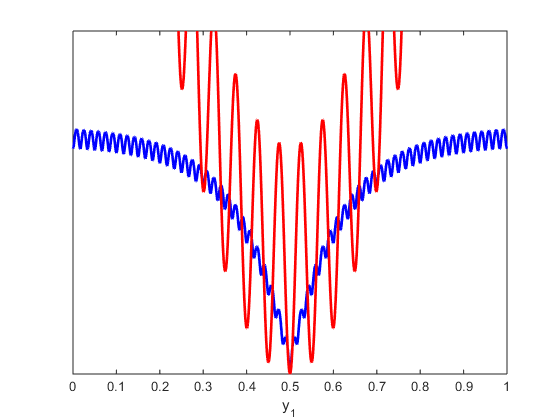}
  \label{fig:CI1}
\end{subfigure}%
\begin{subfigure}{.25\textwidth}
  \centering
  \includegraphics[width=\linewidth]{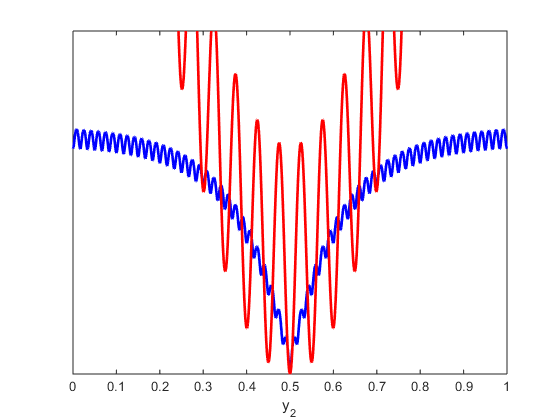}
  \label{fig:CI2}
\end{subfigure}
\caption{In the complete intersection case, all corresponding optimum variables of the two tasks have identical values in the unified search space (see $\boldsymbol{y}_1$ and $\boldsymbol{y}_2$)}
\label{fig:CompleteIntersection}
\end{figure}

For elaboration, the notion of complete intersection of global optima is illustrated in Fig. \ref{fig:CompleteIntersection} when $D=2$. Each figure represents a projection onto the first and the second axis in the unified space, and the global optima of the two tasks are completely intersecting in the unified search space.

\subsubsection{Complete Intersection and High Similarity (CI+HS)}
Task 1 is 50D rotated Griewank, and Task 2 is 50D rotated Rastrigin, which are defined as follows. The global optima of Task 1 and Task 2 are $\boldsymbol{o}_{ch1} = \boldsymbol{o}_{ch2} = (0,0,\dots,0) \in \mathbb{R}^{50}$, and the similarity between them is 1.0000.
\begin{equation}
\begin{split}
Task~1: ~ \min ~&f_1(x) = 1 + \frac{1}{4000} \sum_{i=1}^D z^2_i - \prod_{i=1}^D \cos\Big(\frac{z_i}{\sqrt{i}}\Big), \\
& ~~~\boldsymbol{z} = \boldsymbol{M}_{ch1} \boldsymbol{x}, ~D=50,~ \boldsymbol{x} \in [-100, 100]^D\\
Task~2: ~ \min ~&f_2(x)  = \sum_{i=1}^D \big(z_i^2 -10\cos(2\pi z_i)+10\big), \\
& ~~~\boldsymbol{z} = \boldsymbol{M}_{ch2} \boldsymbol{x}, ~D=50,~\boldsymbol{x} \in [-50, 50]^D
\end{split}
\end{equation}

\subsubsection{Complete Intersection and Medium Similarity (CI+MS)}
Task 1 is 50D rotated Ackley, and Task 2 is 50D rotated Rastrigin, which are defined as follows. The global optima of Task 1 and Task 2 are $\boldsymbol{o}_{cm1} = \boldsymbol{o}_{cm2} = (0,0,\dots,0) \in \mathbb{R}^{50}$. And the similarity between the two tasks is 0.2261.
\begin{equation}
\begin{split}
Task~1: ~ \min ~&f_1(x) = -20 \exp\Bigg(-0.2 \sqrt{\frac{1}{D} \sum_{i=1}^D z_i^2}\Bigg) \\
& ~~ - \exp\Bigg(\frac{1}{D} \sum_{i=1}^D \cos(2\pi z_i)\Bigg) + 20 + e, \\
&~~~\boldsymbol{z} = \boldsymbol{M}_{cm1} \boldsymbol{x}, ~D = 50, ~\boldsymbol{x}\in [-50,50]^D \\
Task~2: ~ \min ~&f_2(x)  = \sum_{i=1}^D \big(z_i^2 -10\cos(2\pi z_i)+10\big), \\
& ~~~\boldsymbol{z} = \boldsymbol{M}_{cm2} \boldsymbol{x}, ~D = 50, ~ \boldsymbol{x} \in [-50,50]^D
\end{split}
\end{equation}

\subsubsection{Complete Intersection and Low Similarity (CI+LS)}
Task 1 is 50D rotated Ackley, and Task 2 is 50D Schwefel, which are defined as follows. The global optimal of Task 1 (search space [-50, 50]) is located at $\boldsymbol{o}_{cl1} = (42.0969,\dots,42.0969) \in \mathbb{R}^{50}$ and the global optimal of Task 2 (search space [-500, 500]) is $\boldsymbol{o}_{cl2} =(420.9687,\dots,420.9687) \in \mathbb{R}^{50}$, so that the global optima of the two tasks are in fact completely intersected in the unified search space (based on the random-key representation scheme). The similarity between the two tasks is 0.0002.
\begin{equation}
\begin{split}
Task~1: ~ \min ~&f_1(x) = -20 \exp\Bigg(-0.2 \sqrt{\frac{1}{D} \sum_{i=1}^D z_i^2}\Bigg)\\
&~~ - \exp\Bigg(\frac{1}{D} \sum_{i=1}^D \cos(2\pi z_i)\Bigg)  + 20 + e, \\
\boldsymbol{z} = &\boldsymbol{M}_{cl1} (\boldsymbol{x} - \boldsymbol{o}_{cl1}), ~D=50, ~\boldsymbol{x} \in [-50,50]^D \\
Task~2: ~ \min ~&f_2(x)  = 418.9829\times D - \sum_{i=1}^D x_i \sin\big(|x_i|^\frac{1}{2}\big), \\
& ~D=50, ~\boldsymbol{x} \in [-500, 500]^D
\end{split}
\end{equation}

\subsection{Benchmark Problems with Partial Intersection of Global Optima}
\begin{figure}
\centering
\begin{subfigure}{.25\textwidth}
  \centering
  \includegraphics[width=\linewidth]{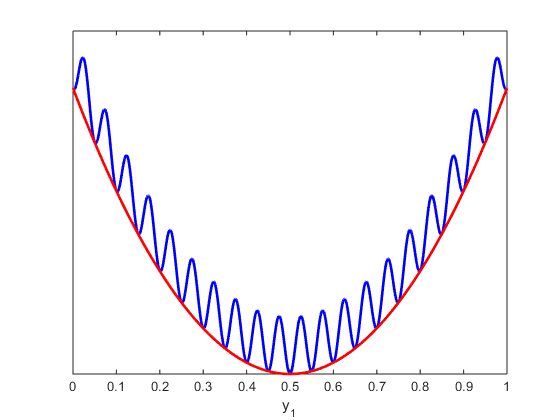}
  \label{fig:PI1}
\end{subfigure}%
\begin{subfigure}{.25\textwidth}
  \centering
  \includegraphics[width=\linewidth]{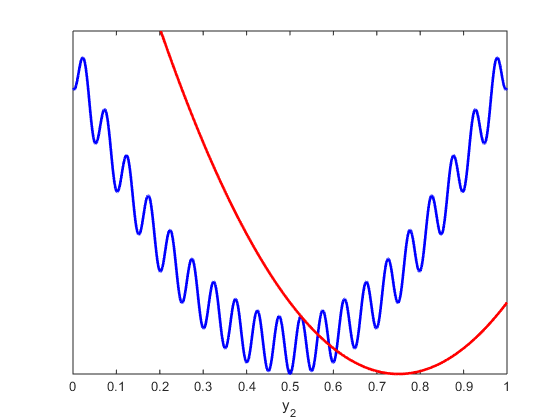}
  \label{fig:PI2}
\end{subfigure}
\caption{In the partial intersection case, some of the corresponding optimum variables (such as $\boldsymbol{y}_1$) for the two tasks have identical values in the unified space. Other optimum variables (such as $\boldsymbol{y}_2$) may have distinct values in the unified space.}
\label{fig:PartialIntersection}
\end{figure}

The notion of partial intersection of global optima is illustrated in Fig. \ref{fig:PartialIntersection} when $D=2$. 

\subsubsection{Partial Intersection and High Similarity (PI+HS)}
Task 1 is 50D rotated Rastrigin, and Task 2 is 50D shifted Sphere, which are defined as follows. The global optima of Task 1 and Task 2 are located at $\boldsymbol{o}_{ph1} = (0,0,\dots,0)\in \mathbb{R}^{50}$ and $\boldsymbol{o}_{ph2} = (0,\dots,0,20,\dots,20) \in \mathbb{R}^{50}$ (first twenty five variables are 0s, and the rest are 20s), respectively. The similarity between the two tasks is 0.8670.
\begin{equation}
\begin{split}
Task~1: ~ \min ~&f_1(x) = \sum_{i=1}^D \big(z_i^2 -10\cos(2\pi z_i)+10\big), \\
&~\boldsymbol{z} = \boldsymbol{M}_{ph1} \boldsymbol{x}, ~D=50, ~\boldsymbol{x} \in [-50,50]^D\\
Task~2: ~ \min ~&f_2(x)  = \sum_{i=1}^D z_i^2, \\
&~\boldsymbol{z} = \boldsymbol{x} - \boldsymbol{o}_{ph2}, ~D=50, ~\boldsymbol{x} \in [-100, 100]^D
\end{split}
\end{equation}

\subsubsection{Partial Intersection and Medium Similarity (PI+MS)}
Task 1 is 50D rotated and shifted Ackley, and Task 2 is 50D Rosenbrock, which are defined as follows. The global optima of Task 1 and Task 2 are located at $\boldsymbol{o}_{pm1} = (0,0,\dots,0, 1,1,\dots,1) \in \mathbb{R}^{50}$ (first twenty five variables are 0s, and the rest are 1s) and $\boldsymbol{o}_{pm2} = (1,1,\dots,1) \in \mathbb{R}^{50}$, respectively. The similarity between the two tasks is 0.2154.
\begin{equation}
\begin{split}
Task~1: ~ \min ~&f_1(x) = -20 \exp\Bigg(-0.2 \sqrt{\frac{1}{D} \sum_{i=1}^D z_i^2}\Bigg) \\
&- \exp\Bigg(\frac{1}{D} \sum_{i=1}^D \cos(2\pi z_i)\Bigg)  + 20 + e, \\
&\boldsymbol{z} = \boldsymbol{M}_{pm1} \boldsymbol{x}, ~D=50,~\boldsymbol{x} \in [-50, 50]^D \\
Task~2: ~ \min ~&f_2(x)  = \sum_{i=1}^{D-1}(100(x_i^2 - x_{i+1})^2 + (x_i - 1)^2),\\
&D = 50, \boldsymbol{x} \in [-50,50]^D\\
\end{split}
\end{equation}

\subsubsection{Partial Intersection and Low Similarity (PI+LS)}
Task 1 is 50D rotated Ackley, and Task 2 is 25D rotated Weierstrass, which are defined as follows. The global optima of Task 1 and Task 2 are located at $\boldsymbol{o}_{pl1} = (0,0,\dots,0) \in \mathbb{R}^{50}$ and $\boldsymbol{o}_{pl2} = (0,0,\dots,0) \in \mathbb{R}^{25}$, respectively. The similarity between the two tasks is 0.0725.
\begin{equation}
\begin{split}
Task~1: ~ \min ~&f_1(x) = -20 \exp\Bigg(-0.2 \sqrt{\frac{1}{D} \sum_{i=1}^D z_i^2}\Bigg) \\
&- \exp\Bigg(\frac{1}{D} \sum_{i=1}^D \cos(2\pi z_i)\Bigg)  + 20 + e, \\
&\boldsymbol{z} = \boldsymbol{M}_{pl1} \boldsymbol{x}, ~D = 50, ~\boldsymbol{x} \in [-50, 50]^D \\
Task~2: ~ \min ~&f_2(x)  = \sum_{i=1}^D \Bigg(\sum_{k=0}^{k_{\max}}\big[a^k \cos\big(2\pi b^k (z_i +0.5)\big)\big]\Bigg)\\
&- D \sum_{k=0}^{k_{\max}} \big[a^k \cos(2\pi b^k \cdot 0.5)\big]\\
&a=0.5,b=3,k_{\max}=20,\\
&\boldsymbol{z} = \boldsymbol{M}_{pl2} \boldsymbol{x}, ~ D = 25, ~ \boldsymbol{x} \in [-0.5,0.5]^D
\end{split}
\end{equation}

\subsection{Benchmark Problems with No Intersection of Global Optima}
The notion of no intersection of global optima is illustrated in Fig. \ref{fig:CompleteIntersection} when $D=2$. From the figures, we see that there is no intersection of the global optima of the two tasks in the unified search space.

\begin{figure}
\centering
\begin{subfigure}{.25\textwidth}
  \centering
  \includegraphics[width=\linewidth]{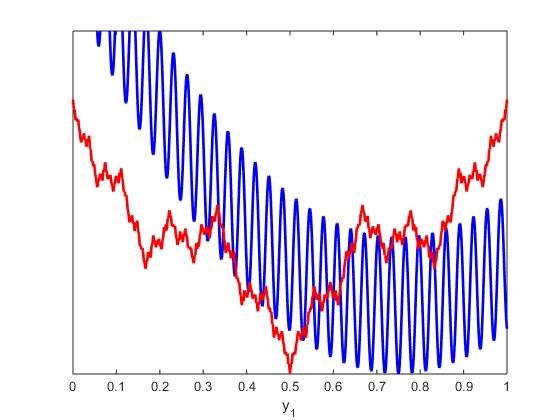}
  \label{fig:NI1}
\end{subfigure}%
\begin{subfigure}{.25\textwidth}
  \centering
  \includegraphics[width=\linewidth]{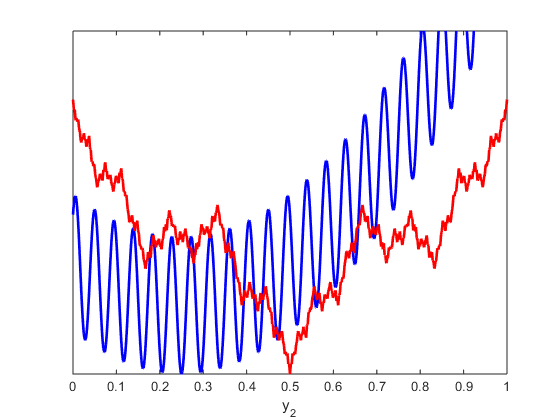}
  \label{fig:NI2}
\end{subfigure}
\caption{In the no intersection case, the global optima of the two tasks are distinct with respect to all variables in the unified search space (see $\boldsymbol{y}_1$ and $\boldsymbol{y}_2$)}
\label{fig:NoIntersection}
\end{figure}

\subsubsection{No Intersection and High Similarity (NI+HS)}
Task 1 is 50D Rosenbrock, and Task 2 is 50D rotated Rastrigin, which are defined as follows. The global optimal of Task 1 and Task 2 are located at $\boldsymbol{o}_{nh1} = (1,1,\dots,1) \in \mathbb{R}^{50}$ and $\boldsymbol{o}_{nh2} = (0,\dots,0) \in \mathbb{R}^{50}$, respectively. The similarity between the two tasks is 0.9434.
\begin{equation}
\begin{split}
Task~1: ~ \min ~&f_1(x)  = \sum_{i=1}^{D-1}(100(x_i^2 - x_{i+1})^2 + (x_i - 1)^2),\\
&~~~D = 50, \boldsymbol{x} \in [-50,50]^D\\
Task~2: ~ \min ~&f_2(x)  = \sum_{i=1}^D \big(z_i^2 -10\cos(2\pi z_i)+10\big), \\
& ~~~\boldsymbol{z} = \boldsymbol{M}_{nh2} \boldsymbol{x}, ~D=50,~\boldsymbol{x} \in [-50, 50]^D
\end{split}
\end{equation}

\subsubsection{No Intersection and Medium Similarity (NI+MS)}
Task 1 is 50D rotated and shifted Griewank, and Task 2 is 50D rotated Weierstrass, which are defined as follows. The global optimal of Task 1 and Task 2 are located at $\boldsymbol{o}_{nm1} = (10,10,\dots,10) \in \mathbb{R}^{50}$ and $\boldsymbol{o}_{nm2} = (0,\dots,0) \in \mathbb{R}^{50}$, respectively. The similarity between the two tasks is 0.3669.
\begin{equation}
\begin{split}
Task~1: ~ \min ~&f_1(x) = 1 + \frac{1}{4000} \sum_{i=1}^D z^2_i - \prod_{i=1}^D \cos\Big(\frac{z_i}{\sqrt{i}}\Big), \\
\boldsymbol{z} = \boldsymbol{M}_{nm1}& (\boldsymbol{x} - \boldsymbol{o}_{nm1}), ~D=50,~ \boldsymbol{x} \in [-100, 100]^D\\
Task~2: ~ \min ~&f_2(x)  = \sum_{i=1}^D \Bigg(\sum_{k=0}^{k_{\max}}\big[a^k \cos\big(2\pi b^k (z_i +0.5)\big)\big]\Bigg)\\
&- D \sum_{k=0}^{k_{\max}} \big[a^k \cos(2\pi b^k \cdot 0.5)\big]\\
&a=0.5,b=3,k_{\max}=20,\\
&\boldsymbol{z} = \boldsymbol{M}_{nm2} \boldsymbol{x}, ~ D = 25, ~ \boldsymbol{x} \in [-0.5,0.5]^D
\end{split}
\end{equation}

\begin{table*}[!t]
\centering
\caption{Average performances (mean and bracketed standard deviation) of MFEA and SOEA on different tasks. Better scores are shown in bold.}
\label{tab:performance}
\begin{tabular}{cccccccc}
\hline
\multirow{2}{*}{Category} & \multicolumn{3}{c}{MFEA} &	&	\multicolumn{3}{c}{SOEA} \\ \cline{2-4} \cline{6-8} & $T_1$ & $T_2$ & score & & $T_1$ & $T_2$ & score \\ \hline
\multirow{2}{*}{CI+HS}	&	\textbf{0.3732}	&	\textbf{194.6774}	&	\multirow{2}{*}{\textbf{-37.6773}}	&	&	0.9084		&	410.3692	&	\multirow{2}{*}{37.6773}	\\	
						&	(0.0617)		&	(34.4953)			&										&	&	(0.0585)	&	(49.0439)	&	\\	\hline
\multirow{2}{*}{CI+MS}	&	\textbf{4.3918}	&	\textbf{227.6537}	&	\multirow{2}{*}{\textbf{-25.2130}}	&	&	5.3211		&	440.5710	&	\multirow{2}{*}{25.2130}	\\	
						&	(0.4481)		&	(52.2778)			&										&	&	(1.2338)	&	(65.0750)	&	\\	\hline
\multirow{2}{*}{CI+LS}	&	\textbf{20.1937}&	\textbf{3700.2443}	&	\multirow{2}{*}{\textbf{-25.7157}}	&	&	21.1666		&	4118.7017	&	\multirow{2}{*}{25.7157}	\\	
						&	(0.0798)		&	(429.1093)			&										&	&	(0.2010)	&	(657.2786)	&	\\	\hline
\multirow{2}{*}{PI+HS}	&	613.7820	&	\textbf{10.1331}	&	\multirow{2}{*}{\textbf{-6.8453}}	&	&	\textbf{445.1040}	&	83.9985	&	\multirow{2}{*}{6.8453}	\\	
						&	(131.0438)		&	(2.4734)			&										&	&	(57.2891)	&	(17.1924)	&	\\	\hline
\multirow{2}{*}{PI+MS}	&	\textbf{3.4988}	&	\textbf{702.5026}	&	\multirow{2}{*}{\textbf{-33.1556}}	&	&	5.0665		&	23956.6394	&	\multirow{2}{*}{33.1556}	\\	
						&	(0.6289)		&	(267.8558)			&										&	&	(0.4417)	&	(10487.2597)&	\\	\hline
\multirow{2}{*}{PI+LS}	&	20.0101&	19.3731	&	\multirow{2}{*}{36.1798}			&	&	\textbf{5.0485}		&	\textbf{13.1894}		&	\multirow{2}{*}{\textbf{-36.1798}}	\\	
						&	(0.1302)		&	(1.7291)			&										&	&	(0.6299)	&	(2.3771)	&	\\	\hline	
\multirow{2}{*}{NI+HS}	&\textbf{1008.1740}	&	\textbf{287.7497}	&	\multirow{2}{*}{\textbf{-33.7021}}	&	&	24250.9184	&	447.9407	&	\multirow{2}{*}{33.7021}	\\	
						&	(346.1264)		&	(92.4182)			&										&	&	(5842.0394)	&	(61.1624)	&	\\	\hline
\multirow{2}{*}{NI+MS}	&	\textbf{0.4183}	&	\textbf{27.1470}	&	\multirow{2}{*}{\textbf{-35.2738}}	&	&	0.9080		&	36.9601		&	\multirow{2}{*}{35.2738}	\\	
						&	(0.0654)		&	(2.6883)			&										&	&	(0.0702)	&	(3.4558)	&	\\	\hline
\multirow{2}{*}{NI+LS}	&	650.8576&	\textbf{3616.0492}	&	\multirow{2}{*}{4.2962}				&	&	\textbf{437.9926}	&	4139.8903	&	\multirow{2}{*}{\textbf{-4.2962}}	\\	
						&	(98.6871)		&	(325.0275)			&										&	&	(62.6339)	&	(524.4335)	&	\\	\hline	
\end{tabular}
\end{table*}

\subsubsection{No Intersection and Low Similarity (NI+LS)}
Task 1 is 50D rotated Rastrigin, and Task 2 is 50D Schwefel, which are defined as follows. The global optimal of Task 1 and Task 2 are located at $\boldsymbol{o}_{nl1} = (0, 0, \dots, 0) \in \mathbb{R}^{50}$ and $\boldsymbol{o}_{nl2} = (420.9687, 420.9687, \dots, 420.9687) \in \mathbb{R}^{50}$, respectively. The similarity between the two tasks is 0.0016.
\begin{equation}
\begin{split}
Task~1: ~ \min ~&f_1(x) = \sum_{i=1}^D \big(z_i^2 -10\cos(2\pi z_i)+10\big), \\
&~\boldsymbol{z} = \boldsymbol{M}_{ph1} \boldsymbol{x}, ~D=50, ~\boldsymbol{x} \in [-50,50]^D\\
Task~2: ~ \min ~&f_2(x)  = 418.9829\times D - \sum_{i=1}^D x_i \sin\big(|x_i|^\frac{1}{2}\big), \\
& ~D=50, ~\boldsymbol{x} \in [-500, 500]^D
\end{split}
\end{equation}

\section{Performance Evaluation and Baseline Results}
In this section, we provide a performance metric to facilitate the comparisons of algorithms by summarizing their overall performance over multiple tasks. 

\subsection{Performance Metric}
In order to compare the performance of different multi- and single-tasking algorithms, a simple performance metric is proposed herein. Say we have $N$ stochastic algorithms, $A_1, A_2, \dots, A_N$ for a specific test case with $K$ minimization tasks $T_1,T_2,\dots,T_K$, and each algorithm is run for $L$ repetitions. Suppose $I(i,j)_l$ denotes the best obtained result on the $l^{th}$ repetition by Algorithm $A_i$ on the task $T_j$. Note that for fairness of comparison, the amount of computational effort (measured in terms of total function evaluations) spent on $T_j$ should be the same for all algorithms (i.e. both single and multitasking). Next, let $\mu_j$ and $\sigma_j$ be the mean and the standard deviation with respect to task $T_j$ over all the repetitions of all algorithms. Thereafter, consider the normalized performance $I'(:,j)_l = (I(:,j)_l-\mu_j)/\sigma_j$. This normalization procedure is repeated for all tasks on all the repetitions. 
 
Based on the above, for each algorithm $A_i$, its final performance score is given as:

\begin{equation}
\mathrm{score}_i = \sum_{j=1}^K \sum_{l=1}^L I'(i,j)_l.
\end{equation}
 
As is clear, a smaller score suggests that the corresponding algorithm has a superior overall performance over all tasks in the multitasking environment. 

\begin{figure*}[!ht]
\centering
\begin{subfigure}{.33\textwidth}
  \centering
  \includegraphics[width=\linewidth]{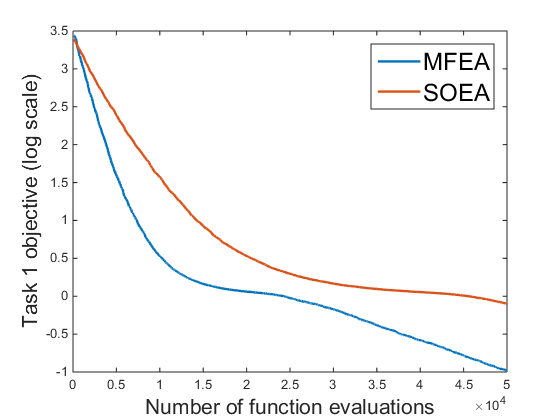}
  \caption{Griewank}
  \label{fig:MFO1Griewank}
\end{subfigure}%
\begin{subfigure}{.33\textwidth}
  \centering
  \includegraphics[width=\linewidth]{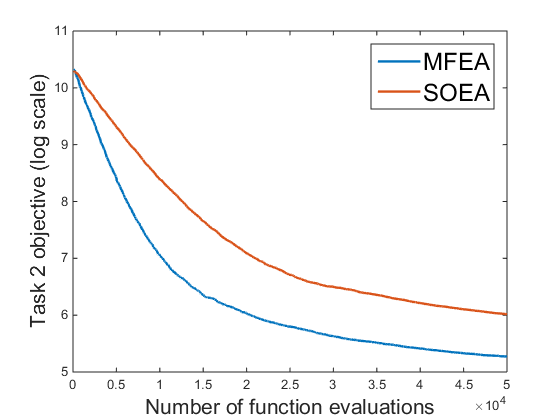}
  \caption{Rastrigin}
  \label{fig:MFO1Rastrigin}
\end{subfigure}
\begin{subfigure}{.33\textwidth}
  \centering
  \includegraphics[width=\linewidth]{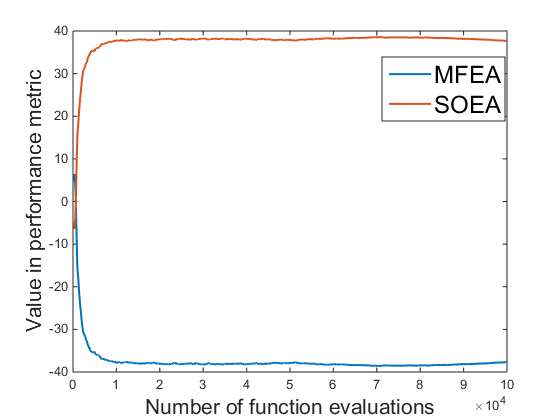}
  \caption{Performance metric trends}
  \label{fig:MFO1metric}
\end{subfigure}
\caption{CIHS: Greiwank and Rastrigin}
\label{fig:MFO1}
\end{figure*}

\begin{figure*}[!ht]
\centering
\begin{subfigure}{.33\textwidth}
  \centering
  \includegraphics[width=\linewidth]{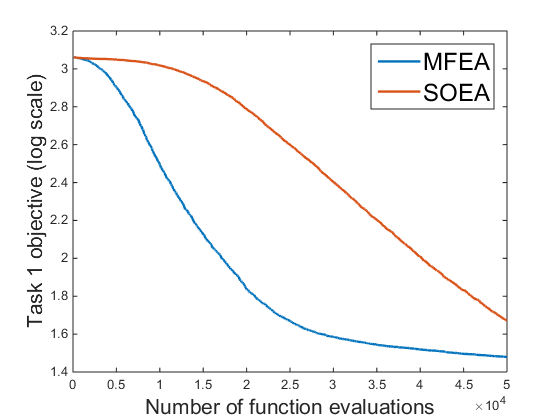}
  \caption{Ackley}
  \label{fig:MFO2Ackley}
\end{subfigure}%
\begin{subfigure}{.33\textwidth}
  \centering
  \includegraphics[width=\linewidth]{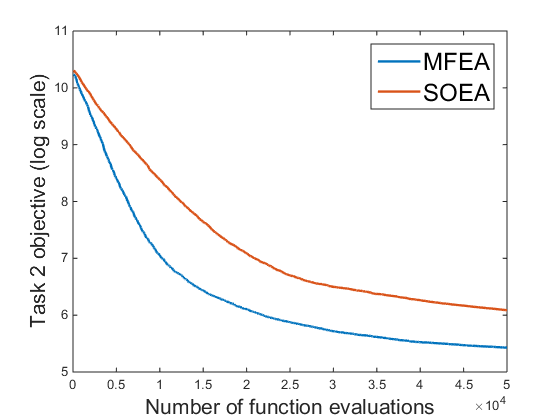}
  \caption{Rastrigin}
  \label{fig:MFO2Rastrigin}
\end{subfigure}
\begin{subfigure}{.33\textwidth}
  \centering
  \includegraphics[width=\linewidth]{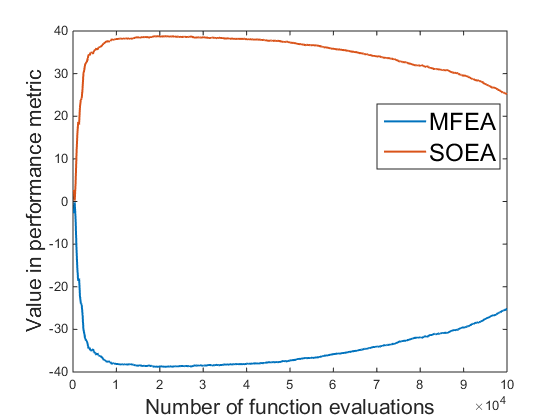}
  \caption{Performance metric trends}
  \label{fig:MFO2metric}
\end{subfigure}
\caption{CIMS: Ackley and Rastrigin}
\label{fig:MFO2}
\end{figure*}

\subsection{Baseline Results}
In Table \ref{tab:performance}, we present the \emph{scores} of the MFEA and SOEA when the termination condition of 100,000 total function evaluations (for both tasks combined) is satisfied. Best scores are shown in bold. Table \ref{tab:performance} demonstrates that multitasking (using MFEA) is in fact superior to single-task optimization in seven out of the nine benchmark examples.

Figs. 9-17 show the sample performances of MFEA and SOEA, where the first two panels are the convergence trends with respect to the two constitutive optimization tasks, and the third panel depicts the ``score" trends of MFEA and SOEA. From these figures as well, it can be inferred that the MFEA demonstrates superior overall search performance in seven out of nine benchmark cases. Note that the two cases where MFEA is inferior, are characterized by very low similarity (Spearman's rank correlation) between the constitutive tasks, thereby highlighting the importance of inter-task synergy to the success of multitasking. The lack of any latent synergy can potentially render multitasking ineffective due to the threat of predominantly negative transfer of genetic material in the unified search space.

\section{Conclusions}
This report presents a total of 9 benchmark problems, and a performance metric, which can be used for future development and testing of multitask optimization algorithms. Baseline results are also provided. As supplementary material, we make available MATLAB implementations of all the toy examples. To conclude, it must be mentioned that the main aim of this work, at least in the long run, is to support the development of an \emph{ideal} evolutionary multitasking engine which is envisioned to be a complex \emph{adaptive} system with performance being comparable (if not consistently superior) to state-of-the-art serial evolutionary optimizers of the present day.

\begin{figure*}
\centering
\begin{subfigure}{.33\textwidth}
  \centering
  \includegraphics[width=\linewidth]{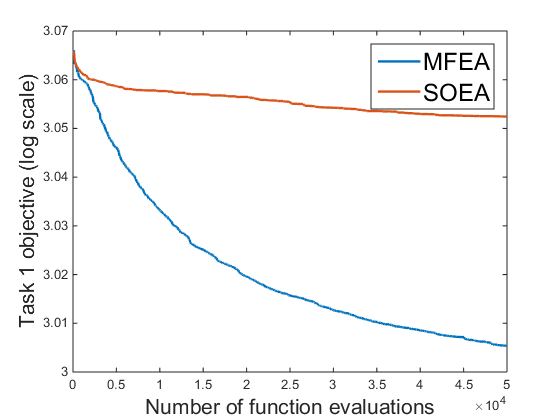}
  \caption{Ackley}
  \label{fig:MFO3Ackley}
\end{subfigure}%
\begin{subfigure}{.33\textwidth}
  \centering
  \includegraphics[width=\linewidth]{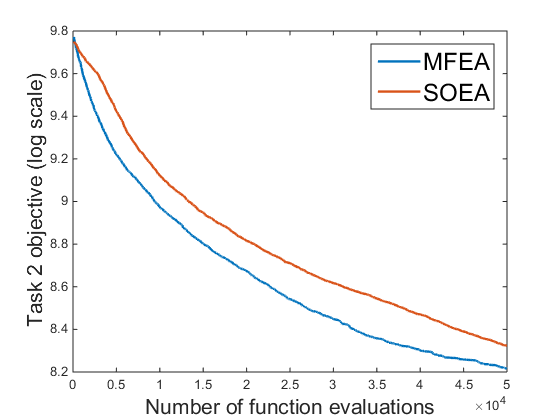}
  \caption{Schwefel}
  \label{fig:MFO3Schwefel}
\end{subfigure}
\begin{subfigure}{.33\textwidth}
  \centering
  \includegraphics[width=\linewidth]{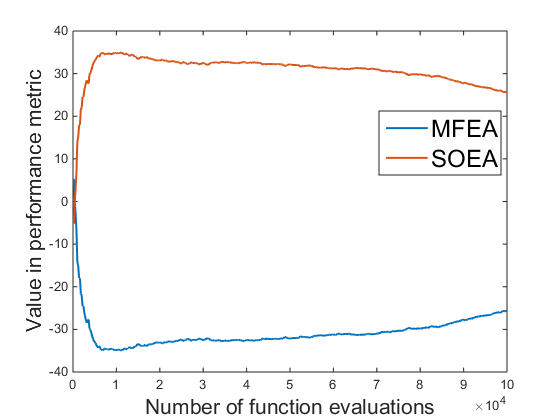}
  \caption{Performance metric trends}
  \label{fig:MFO3metric}
\end{subfigure}
\caption{CILS: Ackley and Schwefel}
\label{fig:MFO3}
\end{figure*}

\begin{figure*}
\centering
\begin{subfigure}{.33\textwidth}
  \centering
  \includegraphics[width=\linewidth]{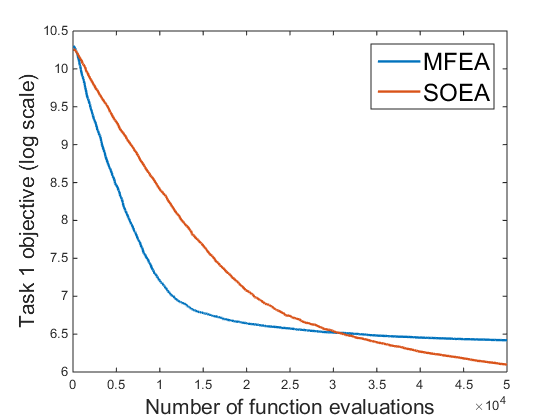}
  \caption{Rastrigin}
  \label{fig:MFO4Rastrigin}
\end{subfigure}%
\begin{subfigure}{.33\textwidth}
  \centering
  \includegraphics[width=\linewidth]{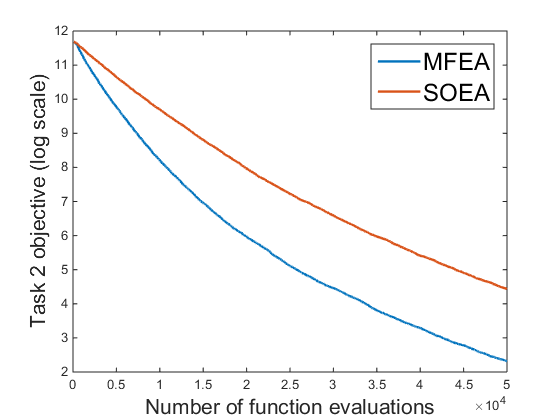}
  \caption{Sphere}
  \label{fig:MFO4Sphere}
\end{subfigure}
\begin{subfigure}{.33\textwidth}
  \centering
  \includegraphics[width=\linewidth]{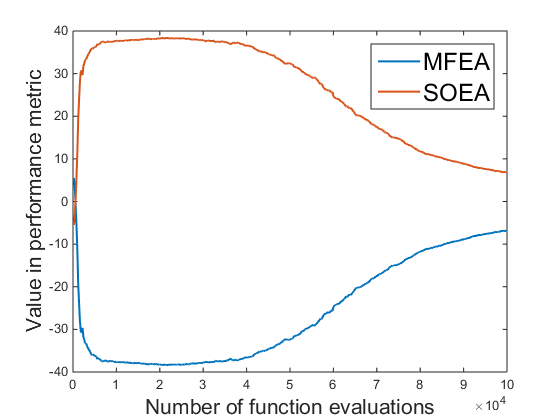}
  \caption{Performance metric trends}
  \label{fig:MFO4metric}
\end{subfigure}
\caption{PIHS: Rastrigin and Sphere}
\label{fig:MFO4}
\end{figure*}

\begin{figure*}
\centering
\begin{subfigure}{.33\textwidth}
  \centering
  \includegraphics[width=\linewidth]{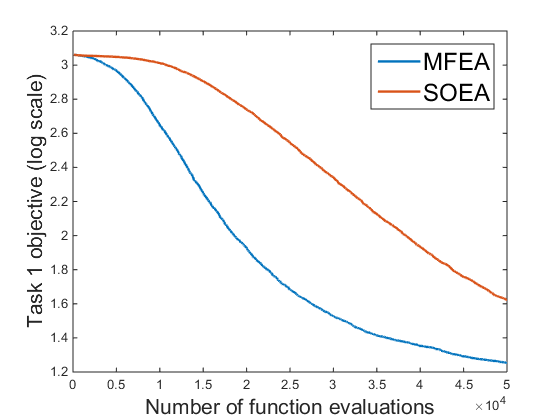}
  \caption{Ackley}
  \label{fig:MFO5Ackley}
\end{subfigure}%
\begin{subfigure}{.33\textwidth}
  \centering
  \includegraphics[width=\linewidth]{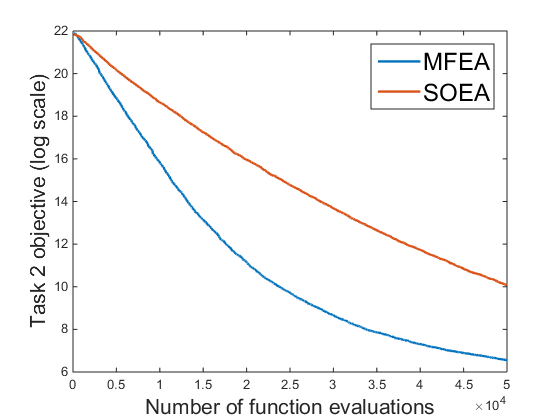}
  \caption{Rosenbrock}
  \label{fig:MFO5Rosenbrock}
\end{subfigure}
\begin{subfigure}{.33\textwidth}
  \centering
  \includegraphics[width=\linewidth]{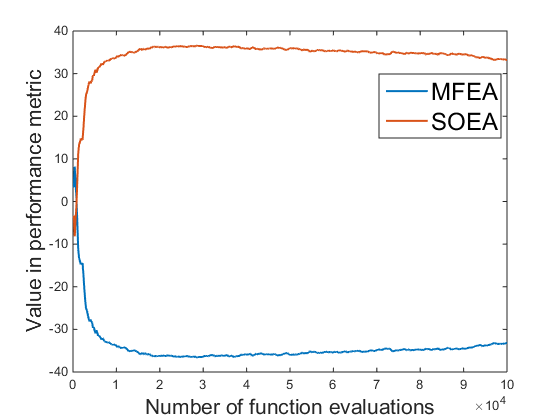}
  \caption{Performance metric trends}
  \label{fig:MFO5metric}
\end{subfigure}
\caption{PIMS: Ackley and Rosenbrock}
\label{fig:MFO5}
\end{figure*}

\begin{figure*}
\centering
\begin{subfigure}{.33\textwidth}
  \centering
  \includegraphics[width=\linewidth]{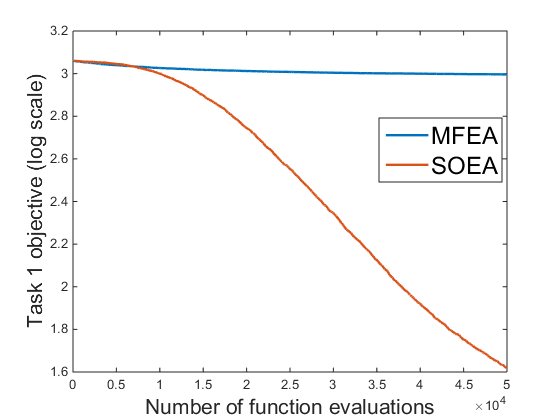}
  \caption{Ackley}
  \label{fig:MFO6Ackley}
\end{subfigure}%
\begin{subfigure}{.33\textwidth}
  \centering
  \includegraphics[width=\linewidth]{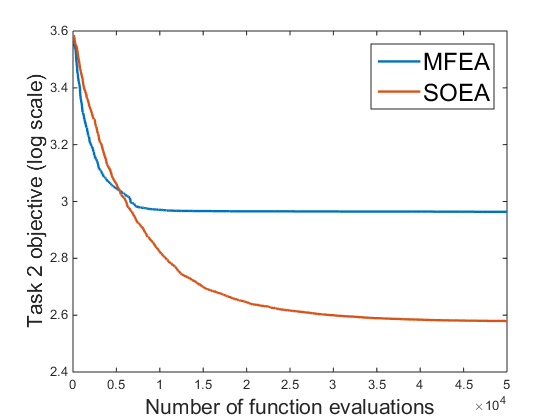}
  \caption{Weierstrass}
  \label{fig:MFO6Weierstrass}
\end{subfigure}
\begin{subfigure}{.33\textwidth}
  \centering
  \includegraphics[width=\linewidth]{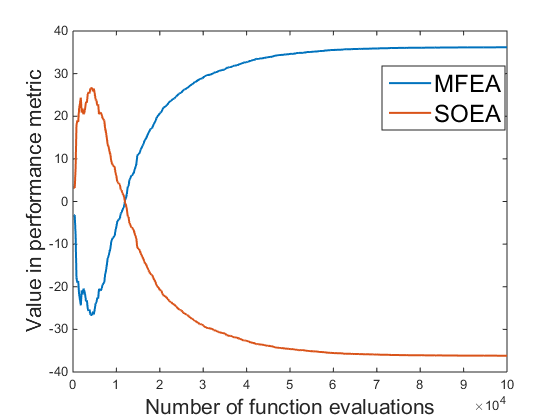}
  \caption{Performance metric trends}
  \label{fig:MFO6metric}
\end{subfigure}
\caption{PILS: Ackley and Weierstrass}
\label{fig:MFO6}
\end{figure*}

\begin{figure*}
\centering
\begin{subfigure}{.33\textwidth}
  \centering
  \includegraphics[width=\linewidth]{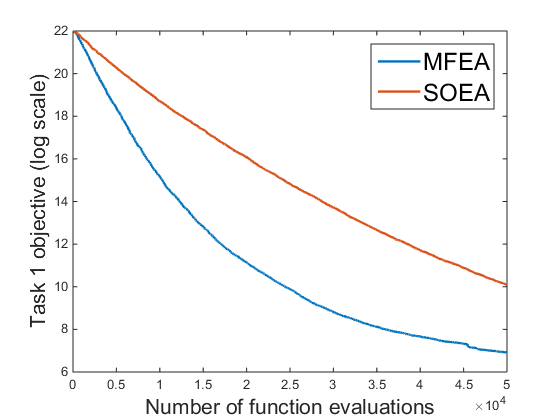}
  \caption{Rosenbrock}
  \label{fig:MFO7Rosenbrock}
\end{subfigure}%
\begin{subfigure}{.33\textwidth}
  \centering
  \includegraphics[width=\linewidth]{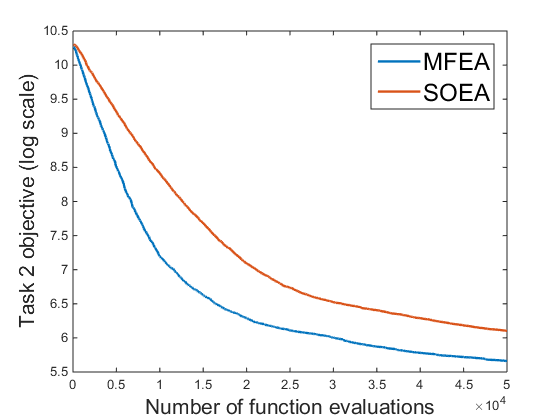}
  \caption{Rastrigin}
  \label{fig:MFO7Rastrigin}
\end{subfigure}
\begin{subfigure}{.33\textwidth}
  \centering
  \includegraphics[width=\linewidth]{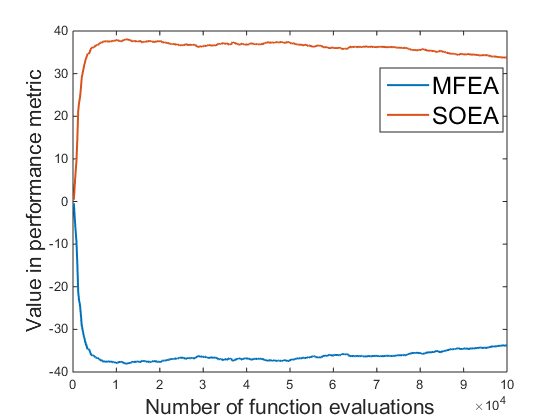}
  \caption{Performance metric trends}
  \label{fig:MFO7metric}
\end{subfigure}
\caption{NIHS: Rosenbrock and Rastrigin}
\label{fig:MFO7}
\end{figure*}

\begin{figure*}
\centering
\begin{subfigure}{.33\textwidth}
  \centering
  \includegraphics[width=\linewidth]{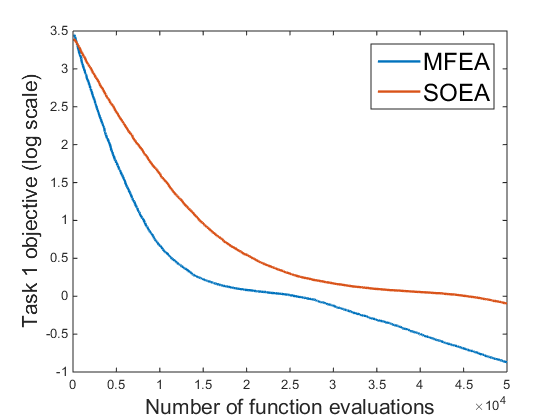}
  \caption{Griewank}
  \label{fig:MFO8Griewank}
\end{subfigure}%
\begin{subfigure}{.33\textwidth}
  \centering
  \includegraphics[width=\linewidth]{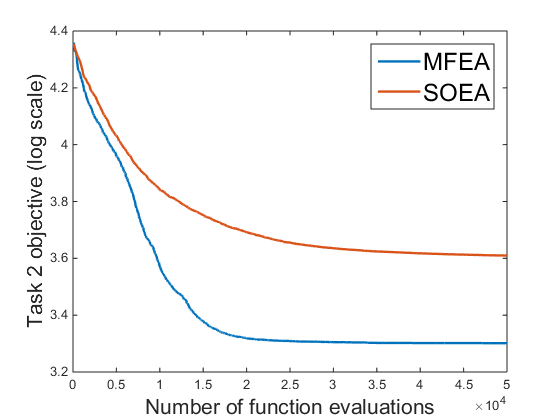}
  \caption{Weierstrass}
  \label{fig:MFO8Weierstrass}
\end{subfigure}
\begin{subfigure}{.33\textwidth}
  \centering
  \includegraphics[width=\linewidth]{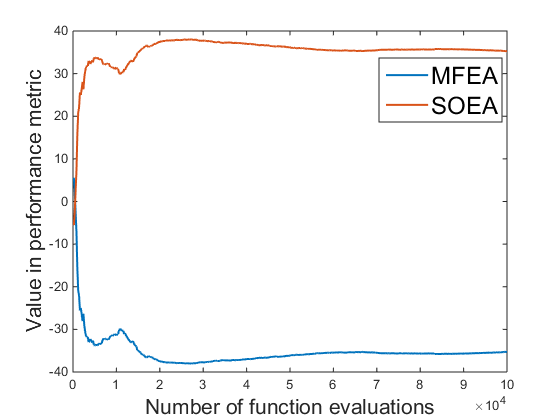}
  \caption{Performance metric trends}
  \label{fig:MFO8metric}
\end{subfigure}
\caption{NIMS: Griewank and Weierstrass}
\label{fig:MFO8}
\end{figure*}

\begin{figure*}[t]
\centering
\begin{subfigure}{.33\textwidth}
  \centering
  \includegraphics[width=\linewidth]{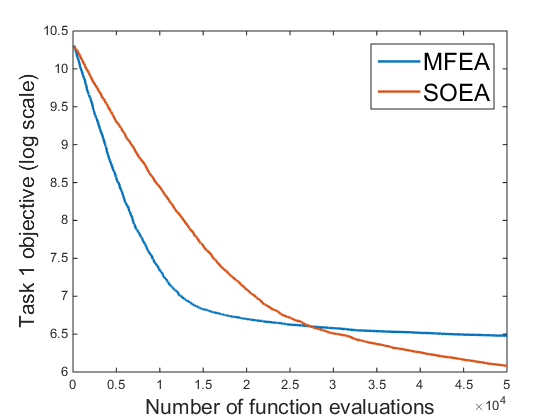}
  \caption{Rastrigin}
  \label{fig:MFO9Rastrigin}
\end{subfigure}%
\begin{subfigure}{.33\textwidth}
  \centering
  \includegraphics[width=\linewidth]{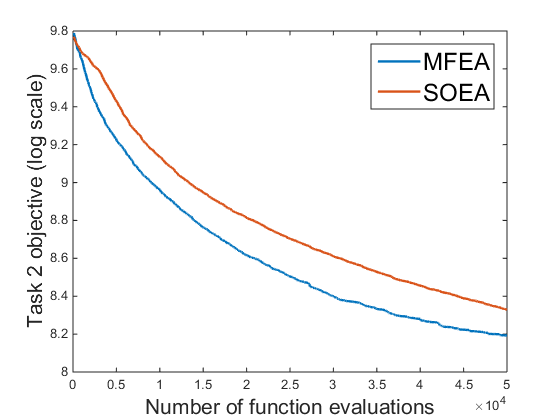}
  \caption{Schwefel}
  \label{fig:MFO9Schwefel}
\end{subfigure}
\begin{subfigure}{.33\textwidth}
  \centering
  \includegraphics[width=\linewidth]{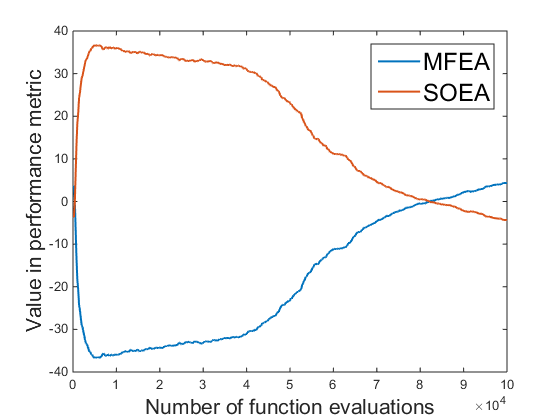}
  \caption{Performance metric trends}
  \label{fig:MFO9metric}
\end{subfigure}
\caption{NILS: Rastrigin and Schwefel}
\label{fig:MFO9}
\end{figure*}

\bibliographystyle{IEEEtran}
\bibliography{references}

\end{document}